\documentclass[10pt,journal,compsoc]{IEEEtran}

\usepackage[utf8]{inputenc} % allow utf-8 input
\usepackage[T1]{fontenc}    % use 8-bit T1 fonts
\usepackage{hyperref}       % hyperlinks
\usepackage{url}            % simple URL typesetting
\usepackage{booktabs}       % professional-quality tables
\usepackage{amsfonts}       % blackboard math symbols
\usepackage{nicefrac}       % compact symbols for 1/2, etc.
\usepackage{microtype}      % microtypography
\usepackage{xcolor}         % colors
\usepackage{amsmath}
\usepackage{graphicx}
\usepackage{subfigure}
\usepackage{url}            % simple URL typesetting
\usepackage{booktabs}       % professional-quality tables
\usepackage{amssymb,xcolor,stackengine,graphicx}
\usepackage{enumitem}
\usepackage{multirow}
\ifCLASSINFOpdf
\else
\fi

\begin{document}
\title{Efficient XAI Techniques: A Taxonomic Survey}

% \tableofcontents
% \newpage

\author{
    Yu-Neng Chuang, Guanchu Wang, Fan Yang, Zirui Liu, \\ Xuanting Cai, Mengnan Du, and Xia Hu
    \IEEEcompsocitemizethanks{\IEEEcompsocthanksitem Yu-Neng Chuang, Guanchu Wang, Fan Yang, Zirui Liu and Xia Hu are with the Department of Computer Science, Rice University. 
    E-mail: \{ynchuang, guanchu.wang, fyang, zl105, xia.hu\}@rice.edu.
    \IEEEcompsocthanksitem Xuanting Cai is with Meta Platforms, Inc.
    Email: \{caixuanting\}@fb.com.
    \IEEEcompsocthanksitem Mengnan Du is with the Department of Data Science, New Jersey Institute of Technology.
    Email: \{mengnan.du\}@njit.edu.
    \IEEEcompsocthanksitem Correspondence to Yu-Neng Chuang and Xia Hu.
    }
}


\markboth{UNDER REVIEW IEEE TRANSACTIONS ON PATTERN ANALYSIS AND MACHINE INTELLIGENCE}%
{Shell \MakeLowercase{\textit{et al.}}: Bare Demo of IEEEtran.cls for IEEE Journals}

\IEEEtitleabstractindextext{
\begin{abstract}
    %Explainable artificial intelligence~(XAI) plays an essential role in the development of trustworthy machine learning systems for real-world problems. 
    Recently, there has been a growing demand for the deployment of Explainable Artificial Intelligence (XAI) algorithms in real-world applications.
    However, traditional XAI methods typically suffer from a high computational complexity problem, which discourages the deployment of real-time systems to meet the time-demanding requirements of real-world scenarios. Although many approaches have been proposed to improve the efficiency of XAI methods, a comprehensive understanding of the achievements and challenges is still needed. To this end, in this paper we provide a review of efficient XAI. 
    Specifically, we categorize existing techniques of XAI acceleration into efficient non-amortized and efficient amortized methods. The efficient non-amortized methods focus on data-centric or model-centric acceleration upon each individual instance. In contrast, amortized methods focus on learning a unified distribution of model explanations,  following the predictive, generative, or reinforcement frameworks, to rapidly derive multiple model explanations. We also analyze the limitations of an efficient XAI pipeline from the perspectives of the training phase, the deployment phase, and the use scenarios. 
    Finally, we summarize the challenges of deploying XAI acceleration methods to real-world scenarios, overcoming the trade-off between faithfulness and efficiency, and the selection of different acceleration methods.
    %The papers studied in this survey are available at \href{https://github.com/ynchuang/awesome-efficient-xai}{{\color{blue}\texttt{https://github.com/ynchuang/awesome-efficient-xai}}}.

\end{abstract}

}

\maketitle
\IEEEpeerreviewmaketitle

\IEEEraisesectionheading{\section{Introduction}}
% \IEEEPARstart{T}{his} 

%XAI in general; 

%XAI needs better efficiency for practical deployment; \\

\IEEEPARstart{M}{achine} learning (ML) has been successfully applied in a variety of domains, such as recommender systems~\cite{he2020lightgcn, chuang2020tpr}, machine translation~\cite{dabre2020survey}, and voice recognition~\cite{chandolikar2022voice}. Despite the advancements in ML, providing transparency in the models, particularly in deep neural networks (DNNs), remains a substantial challenge. The lack of transparency can lead to mistrust and skepticism of ML model predictions, such as the block-box driving decisions made by autopilots. Towards this end, explainable artificial intelligence (XAI) has received increasing attention from the community. A multitude of XAI algorithms have been introduced and can be categorized into two main groups: local and global explanations~\cite{christoph2022xai}. Local explanations aim to explain the reasoning behind an individual model prediction, while global explanations aim to uncover the overall functioning of a complex model by examining its structure and parameters. XAI techniques can also be classified into post-hoc and intrinsic explainability~\cite{du2019techniques}. These XAI methods provide valuable insight into the decision-making process of machine learning models.
In this article, we focus on the efficiency problems of post-hoc local explanation since it is one of the most commonly used methods in the field of XAI. As shown in Figure~\ref{fig:local}, local explanation methods derive feature attribution scores by locally examining the model prediction of each individual data instance one at a time. Traditional post-hoc local explanation methods are particularly inefficient because of the gradual process of creating a unique explainer for each instance. For example, SHAP~\cite{lundberg2017unified} takes around 27 seconds to generate a local model explanation for a $32 \times 32$  CIFAR-10 image.
Considering the efficiency of deriving a model explanation, traditional post-hoc local ones encounter other severe inefficiency issues since each instance requires a unique explainer during the derivation of the explanation.
In addition, the local explanation suffers from extensive computational conditions due to the pending amounts of tested instances, where each instance requires massive permutation times to complete the importance score estimation.
In this manner, two drawbacks push the post-hoc local explanation methods into a difficult situation, obtaining exceptionally high time complexity to fulfill the local explanation requirements.
Despite the high time complexity, the reliable performance still leads to a vast exploitation of post-hoc local methods. Post-hoc local explanation reveals faithful and effective instance-based explanation with theoretical guarantees, such as Shapley-based estimation~\cite{lundberg2017unified} and counterfactual examples~\cite{wachter2017counterfactual}. However, the high computational complexity creates a barrier to deployment in real-world systems with sampling variability~\cite{lundberg2020local}. Providing a real-time explanation is still a remaining challenge in balancing the trade-off between efficacy and efficiency for a post-hoc local explanation method.

Based on the challenges above, we analyze efficiency issues from three different explanation perspectives: individual feature explanations, feature tuple explanations, and influential example explanations. The first class of explanation perspectives focuses on calculating feature scores among all input features, assuming that each feature is mutually independent. Specifically, we discuss the first perspective in the context of feature attribution with acceleration mechanisms. The second perspective involves generating scores for statistical feature interactions. In real-world problem settings, the features are not always independent, which means that the interaction between multiple features may be significantly related to the underlying prediction results.
Unlike the previous two attribution tasks, the last explanation perspective, influential examples, aims to provide an instance that explicitly helps users understand what is happening inside the model. The generated explanation instance serves as a proxy for transparency toward the prediction models, increasing the trustworthiness of the underlying prediction outcomes. We primarily target counterfactual examples (algorithmic recourse) in this taxonomic review as a representative. For simplicity and clarity, we use the terms "counterfactual example" and "algorithmic recourse" interchangeably throughout the rest of this survey.
% This means that the correlations between each feature need to be considered during explanation derivation. In this work, we focus on discussing the efficiency issues of pairwise statistical interaction as a representative of interaction detection. 

We further categorize the efficient approaches based on their acceleration actions, including non-amortized acceleration and amortized acceleration. These two acceleration methods greatly help users to trust the model prediction in a real-time scenario. 
The first line of work locally accelerates the interpretation process for each instance, trying to decrease the computational complexity before or during the explanation generation.
However, non-amortized ones may encounter a limitation while keeping the underlying explanation accuracy intact since original feature information is diminished. The second line of work is amortized acceleration, which utilizes a DNN model to capture the explanation distribution among all training instances globally. The derivation time of amortized explainers can be largely reduced since it provides the explanation via a single feedforward process. However, it may sacrifice explanation performance, since the explainers learn the general explanation distribution instead of focusing on a single data instance. 
Non-amortized acceleration can preserve higher explanation performance while speeding up individual derivation processes, but it is still slower than amortized ones. Amortized acceleration methods can provide a robust and real-time explanation by exploiting one DNN explainer, but they may sacrifice explanation performance.

%In this article, we discuss efficiency issues on post-hoc local explanations from three perspectives. 
The rest of the article is organized into sections as outlined below.
In Section 2, we first introduce the context and formulation of efficiency issues to post-hoc local explanation methods. We then summarize the current state-of-the-art in efficient explanation methods, including non-amortized acceleration in Section 3 and amortized acceleration in Section 4. Following that, in Sections 5 and 6, we present challenges and future directions for improving the efficiency of explanation derivation. Finally, in Section 7, we summarize the discussions in this work.

\section{Background of Efficient Explainable Artificial Intelligence}
This section depicts the efficiency issues that explainable artificial intelligence (XAI) faces, as well as the requirements for accelerating strategies to address efficiency concerns in explanation tasks.

\subsection{Efficient Issues on XAI}

\begin{figure}[t]
    \centering
    \includegraphics[width=0.4\textwidth]{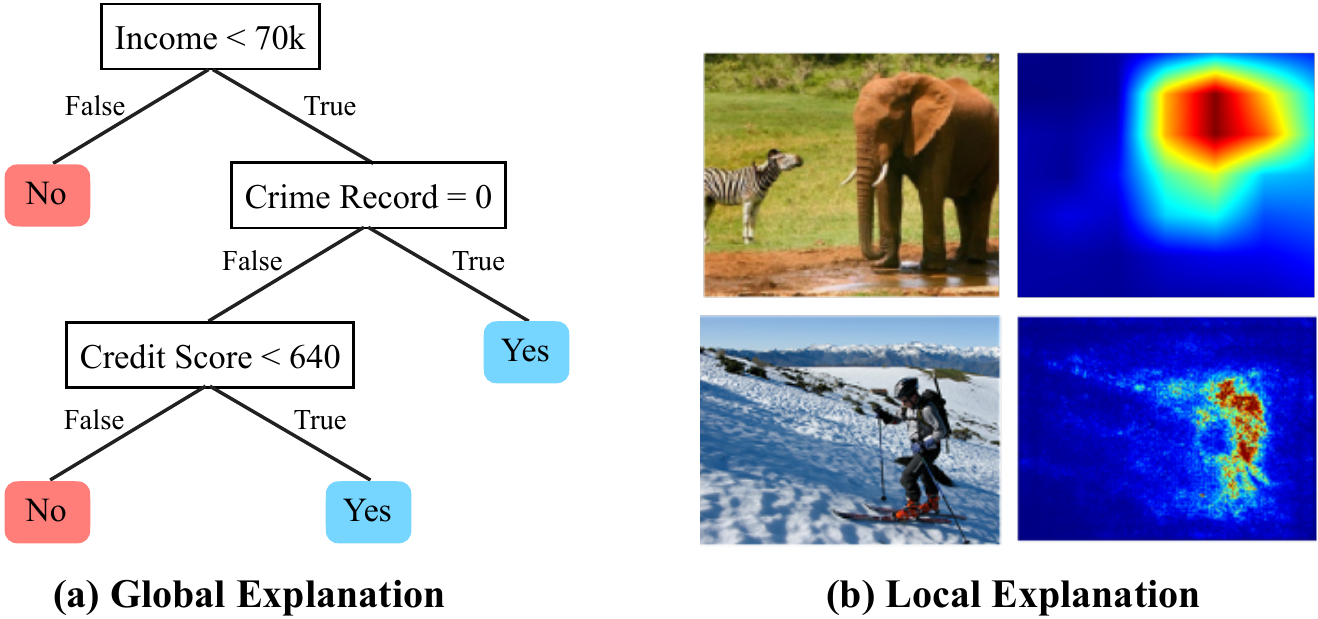}
    \vspace{-0.1cm}
    \caption{Comparison between global (model-wise) explanation and local (instance-wise) explanation. (a) The decision tree intrinsically illustrates the model-wise important features of loan decisions from banks; (b) while heatmaps produced the instance-wise feature attributions toward each individual on the image classification task. In this survey, we focus on the efficient issues in local explanation.}
    \label{fig:local}
\end{figure}

Post-hoc local explanation is the most widely studied method that aims to provide instance-based explanations to locally examine model predictions (see Figure~\ref{fig:local})~\cite{du2019techniques}. This paradigm has been widely used in real-world systems due to its effectiveness and faithfulness. Moreover, some previous work~\cite{slack2021reliable,agarwal2022rethinking} focus on eliminating the uncertainty of post-hoc local explanations to derive more stable, consistent, and reliable model explanations, which can increase user trust in the decisions made by prediction models.
However, some traditionally local explanation techniques, such as Shapley-based methods~\cite{lundberg2017unified, shap_inter_idx}, are seriously suffering from efficiency issues, although their generated model explanations are faithfulness and theoretical guarantee.
% However, post-hoc local explanations have a major drawback in terms of efficiency, as they rely heavily on large amounts of permutation and need to go through all queried instances one by one. 
As the experimental results compared to GT-Shapley (ground truth Shapley value) shown in Table~\ref{tab:ks_sample}, traditional XAI requires nearly 1.0 seconds to obtain a satisfactory explanation result under 4,000 permutation times. In contrast, efficient XAI only needs 0.00015 seconds to complete the explanation derivation process, which is much faster than GT-Shapley and traditional XAI. In the following sections, we will discuss the challenges and limitations of post-hoc local explanations in the context of three different tasks: feature attribution, statistical interaction detection, and counterfactual examples.

\begin{table}[t]
\centering
\caption{Experiment of feature attribution task on Adult Dataset(13 features). The experiments compare the deriving time with Ground-truth Shapley Values(GT-Shapley), Traditional XAI (SHAP~\cite{lundberg2017unified}), and Efficient XAI (SHEAR \& CoRTX). SHAP-X represents SHAP with X-times of permutation. "Second" (sec.) is the measurement unit of deployment time per instance.}
% "Second" is the unit of measuring training time per data instance, while 
\setlength{\tabcolsep}{4pt}{
\begin{tabular}{l | ccc ccc ccc ccc ccc ccc |}
\toprule
& GT-Shapley & SHAP-4000 & SHEAR~\cite{wang2022accelerating} & CoRTX~\cite{cortx} \\
\midrule\midrule 
 % & & 41.6142 & --- & --- & --- & --- & --- \\
$\boldsymbol\ell_2\text{-error}$ & -- & 0.0021 & 0.0019 & 0.007 \\
Time/instance & 5.8536 & 0.9869 & 0.0141 & 0.00015\\
\bottomrule
\end{tabular}}
\label{tab:ks_sample}%
\end{table}%

\begin{itemize}[leftmargin=*]
%\vspace{0.7mm}
% \noindent
\item \textbf{Feature Attribution Tasks:}
In feature attribution tasks, the goal of efficient XAI is to accelerate the generation of feature importance scores for a model explanation. In this paper, we focus on Shapley value-based feature attribution tasks, which have been shown to be inefficient.
Shapley values~\cite{kuhn1953contributions} originally aimed to estimate the feature contribution in the cooperative game theory. In feature attribution tasks, they are often used as the important scores of imputing feature set $\mathcal{U} = \{ 1, \cdots, M \}$ to the black-box model behaviors. For any value function $f: 2^M \rightarrow \mathbb{R}$, the Shapley values $\phi_i (f, \mathcal{U}) \in \mathbb{R}$ of feature $i$ can be formalized as follows:
\begin{equation}
    \vspace{0.1cm}
    \phi_i (f, \mathcal{U}) \!=\! \!\! \sum_{\boldsymbol{S} \subseteq \mathcal{U} \setminus \{ i \}} \!\! \binom{M \!-\!\! 1}{|\boldsymbol{S}|}^{-1} \!\!\!\!\! \big[ f(\{ i \} \cup \boldsymbol{S}) \!-\! f(\boldsymbol{S}) \big].
    \label{eq:shapley_value}
    \vspace{0.1cm}
\end{equation}
where $S \subseteq \mathcal{U}$ is a feature coalition set. In other words, the average preceding difference considering all possible feature coalitions indicates the contribution of feature $i$. Nevertheless, the calculation of Shapley values relies on $2^M$ times of model evaluation to estimate the contribution of feature $i$, where the computational complexity is $T [ \phi_i (f, \mathcal{U}) ] = O(2^M)$. One of the inefficient challenges is how to rapidly estimate the Shapley values but without traversing through all feature coalitions.
%\vspace{0.7mm}
% \noindent
\item \textbf{Statistical Interaction Detection Tasks:}
Unlike most feature attribution tasks, which assume that features are independent, this task assumes that different features interact with one another. Following the definition of statistical interaction, $I \subseteq \mathcal{U}$ is said to be a feature interaction if and only if there does not exist $f_{\setminus i}$ for $i \in I$ satisfying the following: 
\begin{equation}
    f(\mathcal{U}) = \sum_{i \in I} f_{\setminus i}({\mathcal{U} \setminus \{ i \}})
    \label{eq:inter_def}
\end{equation}
where $f_{\setminus i}$ denotes a function that does not depend on the feature $i$.
To analyze the interaction of features that contributed to the prediction results of the model, the measurement of interaction scores $\phi_I (f, \mathcal{U})$ can be formally defined as follows:
\begin{equation}
    \vspace{0.1cm}
    \phi_I (f, \mathcal{U}) \!=\!\! \sum_{L \subseteq I} (-1)^{|L| - 1}  f(L),
    \label{eq:inter_value}
\end{equation}
where $L$ represents a $n$-variable sub-interaction from $I$. The high-level idea of the definition is to accurately consider all possible sub-interactions to form the actual importance score of feature interaction $I$. However, the computational process is extremely time-consuming and needs to go through all the possible sub-interactions and feature coalitions. Specifically, traditional interaction detection methods require calculating all statistical feature interactions to generate scores among feature interactions, where the number of interaction candidates is
extremely large. The number of interaction candidates is $2^N$ if there exist $N$ features, which makes it not applicable to real-world systems. The goal here is to alleviate the inefficiency of estimating scores for detecting feature interactions. 
%\vspace{0.7mm}
% \noindent
\item \textbf{Counterfactual Example Tasks:}
Another goal is to improve the efficiency challenges in counterfactual explanation tasks. Counterfactual explanations formalize the exploration of “what-if” scenarios, which are an instance of example-based reasoning using a set of hypothetical data samples. Considering a classification model $f: \mathbb{R}^M \rightarrow \{-1, 1\}$ as an example, well-established counterfactuals $x^*$ are required to flip the prediction outcomes of the original queried instance $q_0$. Given $f(q_0) = -1$, the explanation problem can be formally illustrated as follows:
\begin{equation}
    x^* = \arg\min_{x \sim \mathcal{X}} \mathcal{L}(x, q_0) ~~~~\textrm{s.t.}~~ f(q_0) = -1, f(x^*) = 1
    \label{eq:count}
\end{equation}
where $\mathcal{X}$ denotes the potential data distribution to the counterfactual universe of a given queried instance $q_0$, and $\mathcal{L}$ is the distance measurement in the input space. To solve Equation~\ref{eq:count}, traditional post-hoc local ones generate counterfactual examples by gradually adjusting the features to achieve good validity and sparsity~\cite{verma2020counterfactual}.
Under good conditions of validity and sparsity, actionability and closeness are two more essential factors that lead the counterfactual examples to be user accessible. To meet these four requirements, additional operations such as mix-integer programming are necessary to generate counterfactual examples, which results in slow progress and hinders the ability to obtain efficient results.
\end{itemize}

\begin{figure}[t]
    \centering
    \includegraphics[width=0.48\textwidth]{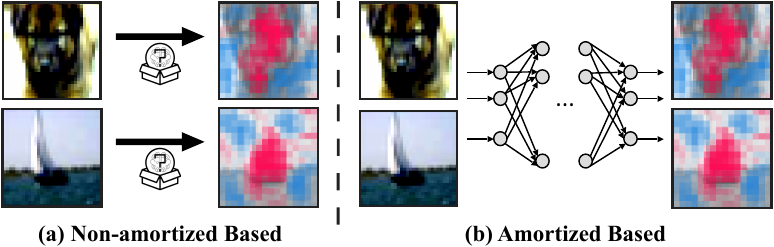}
    \caption{A non-amortized explanation method using four explainers for four instances, and an amortized explanation method exploiting only one explainer among four instances.}
    \label{fig:amort}
\end{figure}

In this paper, we concentrate on addressing the efficiency issues related to the three aforementioned explanation tasks. The goal of efficient XAI is to speed up the generating process while maintaining the effective model explanation and running it in a short period of time. To cope with the challenges, existing work proposes two kinds of framework for efficient XAI: non-amortized acceleration methods and amortized acceleration methods, which are shown in Figure~\ref{fig:amort}.

\subsection{Non-amortized Acceleration Methodology}
Efficient non-amortized methodologies aim to address the efficiency issues of post-hoc local explanation methods on feature attribution, interaction detection, and counterfactual explanation tasks. Non-amortized post-hoc local methodologies require a unique explainer for each instance in order to derive a model explanation, which makes the time complexity grow with the number of tested instances. Despite the high latency in deriving explanations, non-amortized methods are effective at providing satisfactory model explanations because each explainer is only trained on a specific tested instance. Existing approaches can be divided into two groups based on their explanation angles: attribution-oriented and counterfactual-oriented methods.

\subsubsection{Non-amortized Attribution-oriented Methods}
Non-amortized attribution-oriented methods focus on generating instance-wise model explanations, which provides the model explanation of underlying model prediction among each tested feature or feature interaction. 
According to the training paradigms, attribution-oriented methods can be divided into three aspects to accelerate inefficient challenges in Equation (\ref{eq:shapley_value}) and Equation (\ref{eq:inter_value}). The first line of methods~\cite{tsang2020does,covert2021improving} utilizes proxy models (e.g., linear regression) to fit the distribution of feature importance scores and treats the coefficient of proxy models as estimated Shapley values.
The second group of work adopts the preceding difference of the value function to generate the explanation, which averagely calculates the difference in model prediction as the feature importance scores. For example, RISE~\cite{petsiuk2018rise}, Permutation Sampling~\cite{castro2009polynomial}, and Antithetical Permutation Sampling~\cite{mitchell2022sampling} are three representative works. The third group of methods exploits the model gradient to yield a model explanation, such as Integrated Gradient~\cite{sundararajan2017axiomatic}, GradCAM~\cite{selvaraju2017grad} and SmoothGrad~\cite{smilkov2017smoothgrad}. 
Generally, the third group obtains a relatively faster explanation process than the first and second groups. However, extra time is still required to perform the sampling process on each instance while generating the model explanation. The time complexity of these three groups of methods is still highly dependent on the number of sampling and tested instances. Despite their effectiveness and efficacy, attribution-based explanation methods often face the challenge of long computation times, which presents a significant obstacle in deploying non-amortized methods to real-time systems.

\subsubsection{Non-amortized Counterfactual-oriented Methods}
Generally, Equation~(\ref{eq:count}) can be merely solved by algorithm-based counterfactual approaches~\cite{ustun2019actionable,karimi2020algorithmic,karimi2021algorithmic}, which employ different optimization strategies to generate counterfactual examples for each queried instance $q_0$. However, this approach suffers from efficiency issues since the counterfactual instances are decided instance by instance. The optimization process of each new query requires one specific optimization problem of Equation~(\ref{eq:count}) at one time. The explanation progress among a group of queried instances could be exceptionally time consuming.
In this case, existing work provides data adjustment on imputing instances before optimizing the counterfactual distribution,  accelerating the convergence of the training process. Another pack of prior achievements incorporates counterfactual-oriented constraints to speed up the training of yielding counterfactual examples. Despite the efficacy of algorithm-based ones, the high computational cost eventually creates a barrier to deployment on real-world systems that require low latency to provide real-time service.     

\subsection{Amortized Acceleration Methodology}
Unlike non-amortized methods, amortized methodologies utilize a
unified explainer to learn the distribution of model explanation. The goal of amortized ones is to replace the local training pattern in post-hoc local explanation with a deep neural network (DNN). The alternative replacement accelerates the process of generating an instance-based model explanation.
The unified explainer only requires conducting a single feed-forward process in the inference phase. In this way, the time complexity of amortized methods is restricted to constant, which explicitly solves the efficient issues discussed in Section 2.1. Existing methods are designed to deal with two discussed explanation tasks: the feature attribution task and the counterfactual example task. 

\subsubsection{Amortized Attribution-oriented Methods}
Amortized attribution-oriented methods employ predictive DNN models to simultaneously approximate the unified distribution of feature attributions for each instance. The approximation of data distribution has been widely exploited in various domains, such as matrix factorization~\cite{mf,fism} in recommendation systems, to learn the distribution of task-specific information. The deductive output of attribution-oriented models can help humans understand what prediction models consider when making predictions. Amortized methods fundamentally address the efficiency issues of non-amortized methods, which reduce the number of explainers during local explanation derivation. However, finding a balance between explanation effectiveness and efficiency poses a challenge in establishing an optimal training paradigm.

\subsubsection{Amortized Counterfactual-oriented Methods}
Amortized counterfactual-oriented methods use generative-based models or reinforcement agents to learn the general rules of counterfactual examples by solving Equation~(\ref{eq:count}). Different from attribution-oriented ways, which are plausible to rely on ground-truth explanation information, counterfactual cases usually do not obtain ground truth as the label reference when updating the explanation models. The example generation process in the inference phase thereby requires only a single forward passing, making it a real-time explanation derivation.

%\vspace{0.7mm}
\begin{itemize}[leftmargin=*]
% \noindent
 \item \textbf{Generative-based Methods:}
To cope with the challenges mentioned above, counterfactual-oriented methods utilize generative models to synthesize general rules for producing counterfactual examples in the latent code space. Specifically, the generative-based model builds an adversarial learning framework, including a generator to produce counterfactual examples and a discriminator to prevent deviated examples from the generator. The framework typically employs a model based on generative adversarial networks (GANs)~\cite{goodfellow2020generative,mirza2014conditional} and incorporates counterfactual-oriented objectives in the training phase. In this way, generative-based frameworks can accelerate the derivation process by producing multiple counterfactual examples for a given instance simultaneously.
%\vspace{0.7mm}
% \noindent
 \item \textbf{Reinforcement-based Methods:}
Counterfactual-oriented methods take advantage of deep reinforcement agents to efficiently formulate the decision policy to generate counterfactual examples. The decision process for counterfactual examples is typically a discrete action, which is not differentiable and cannot be directly used in model training. Instead, deep reinforcement learning (Deep RL)~\cite{hausknecht2016half,masson2016reinforcement,wei2018hierarchical} employs an action space module to transform discrete actions into corresponding continuous parameters. By taking advantage of the properties of Deep RL, reinforcement agents in counterfactual-oriented methods can synthesize the discrete rules for generating counterfactual examples. The Deep RL framework for model explanation can then accelerate the explanation-generation process with the aid of pre-trained agents during the inference phase.
\end{itemize}

\begin{figure*}[t]
    \centering
    \includegraphics[width=0.9\textwidth]{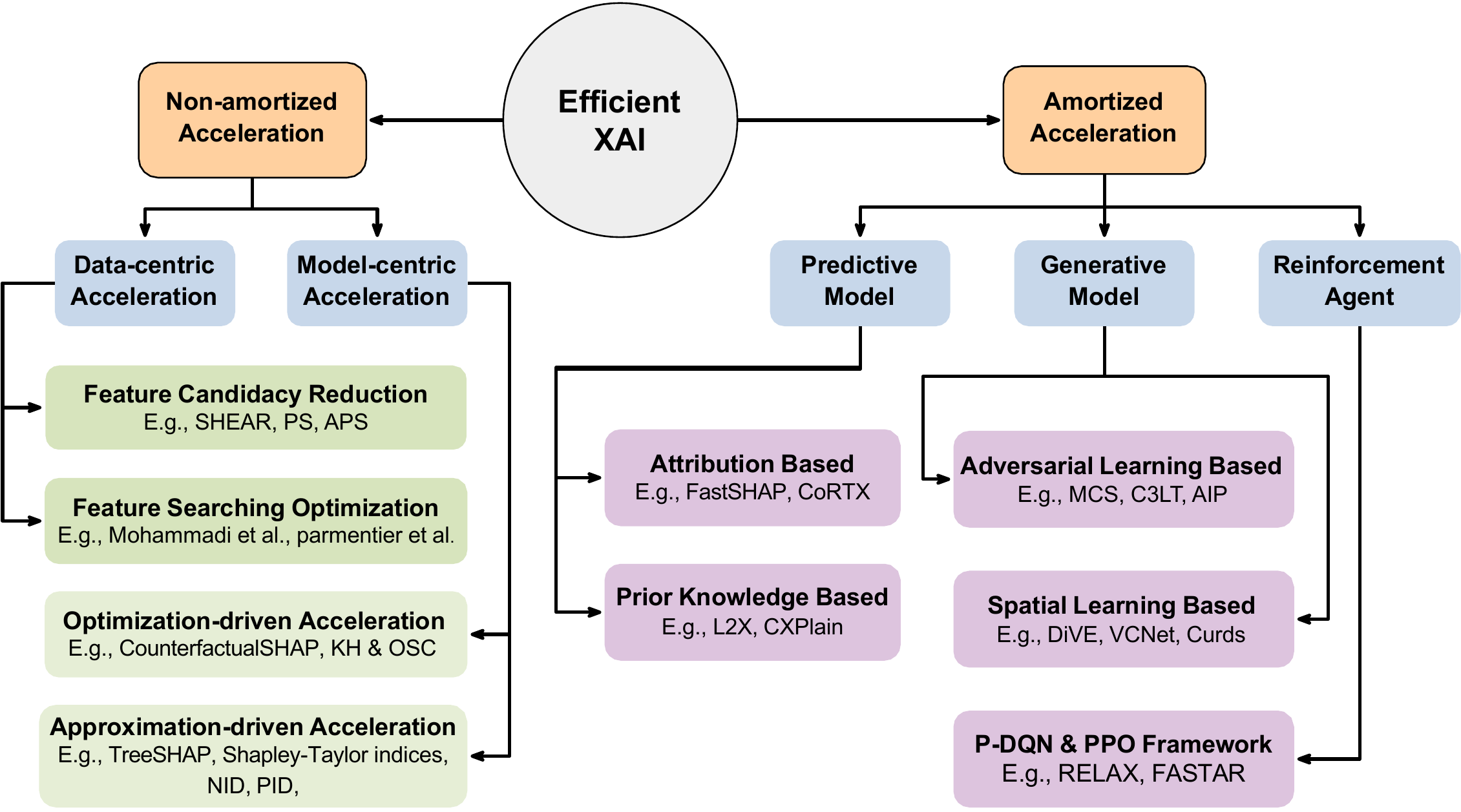}
    \vspace{-0.1cm}
    \caption{Structure of the survey. Two main categorizations are discussed in this paper: Non-amortized Acceleration and Amortized Acceleration.}
    \label{fig:structure-servey}
\end{figure*}

\section{Non-Amortized Acceleration} 
In this section, we divide the non-amortized acceleration into two broad families: data-centric acceleration and model-centric acceleration.

\subsection{Data-centric Acceleration}
The data-centric mechanism adjusts the data to fundamentally improve the performance of the ML model. Superior data quality usually leads to fast convergence of the learning process and competitive prediction results since it eliminates the noise from the original data. The goal of data-centric acceleration is to dismiss the unimportant features or feature coalitions before the head, which reduces the computational complexity of explanation derivation. In Section 3.1.1, we summarize acceleration methods for feature attribution, and in Section 3.1.2, we summarize acceleration methods for counterfactual example generation.

\subsubsection{Feature Candidacy Reduction}
Feature reduction accelerates the estimation of model explanation by eliminating the feature candidates, making the precise computation feasible and practical. The chosen features are typically decided by the causative observations from the prediction models and feature properties, such as pre-calculating contributions of feature that are independent with prediction model~\cite{yang2021fast}. 

One of the existing work SHEAR~\cite{wang2022accelerating} accelerates the estimation of Shapley value by contributive cooperative feature selection. The cooperative features are selected based on the well-contributed feature interactions from the prediction model. The Shapley values are estimated by only considering the selected coalitions of contributive features. Different from SHEAR, another work groupShapley~\cite{jullum2021groupshapley} estimates the Shapley values under feature groups instead of individual features. The feature groups are decided in terms of feature type or dependence. Calculating the Shapley values of given features typically requires iterating all the features in combination with the given features, which inherently increases the execution time. One previous work~\cite{chen2018shapley} also proposes two measures for instance-wise feature importance scoring: L-Shapley and C-Shapley on the graph-structured dataset, which selects the feature candidacy by neighborhood information of the given graph structural dataset. The two measurements compute the Shapley values by exploiting the underlying graph structure, which reduces the time complexity from exponential to linear level.
To cope with the challenge, the target features are replaced with a group of features on deriving the Shapley values, where the time complexity can be decreased once the group number is small.

Considering the intractable situation of lacking prior information, some previous works~\cite{ancona2019explaining,ghorbani2020distributional,bhatt2020evaluating} focus on reducing the number of features by randomly sampling partial candidate coalitions before obtaining feature attribution scores.
The deduction on coalitions usage can directly speed up the calculation process. One of the representative works is permutation sampling~\cite{castro2009polynomial}, which is an unbiased sampling method to speed up the estimation of the Shapley values. As the random sampling is still a bottleneck of efficiency, the antithetic sampling~\cite{lomeli2018antithetic} is combined with the permutation sampling to alleviate the efficient drawback from random sampling.
Specifically, the antithetic sampling method takes the correlated pairs sampling instead of standard i.i.d. random instance sampling. For simplicity in practice, the correlated pairs are considered the complementary pairs of the feature sets.
This enables the pairs of permutations to be negatively correlated, theoretically reducing the estimation variance. Moreover, antithetic sampling reduces half of the sampling time since it only needs to sample one element in each complementary pair. As an efficient and effective method, antithetic permutation sampling has been widely deployed in real-world systems.

\begin{table*}[t!]
    \caption{Assessment of the collected efficient XAI papers on the key properties. "Non-amtz." and "Amtz." represent non-amortized acceleration and amortized acceleration, respectively.
    Details about the methods in the full table are given in Sections 3 and 4. "NLP" represents natural language data, "CV" denotes image data, and "Tab." means tabular data.}
    \centering
    \setlength{\tabcolsep}{5pt}{
	\begin{tabular}{c| c| ccc | cc }
        \toprule
        &\multicolumn{1}{c}{} & \multicolumn{3}{|c}{Development Process} & \multicolumn{2}{|c}{Derivation Process}\\
        \cmidrule(lr){3-5} \cmidrule(lr){6-7}
        & Paper & Modeling Paradigm & Acceleration Methods & Datatype & Shapley Based & Explanation Perspectives \\
        \midrule\midrule
        Non-amtz. & \cite{wang2022accelerating} & Coalition & Sample by Prior Information & Tab. & Yes & Feature Attribution\\
        & \cite{jullum2021groupshapley} & Feature Cluster & Sample by Prior Information & Tab. & Yes & Feature Attribution \\
        & \cite{chen2018shapley} & Graph Neighborhood & Sampling by Prior Information & CV, NLP & Yes & Feature Attribution \\
        & \cite{castro2009polynomial} & Coalition & Random Sampling & Tab. & Yes & Feature Attribution \\
        & \cite{mitchell2022sampling} & Coalition & Antithetic Sampling & CV, Tab. & Yes & Feature Attribution \\
        & \cite{mitchell2022sampling} & Coalition & KH \& OSC & CV, Tab. & Yes & Feature Attribution \\
        & \cite{russell2019efficient} & MIP & Constraint Search & NLP & No & Counterfactuals \\
        & \cite{mohammadi2021scaling} & MIP & Nearest Search & Tab. & No & Counterfactuals \\
        & \cite{parmentier2021optimal} & MIP & Constraint Search & Tab. & No & Counterfactuals \\
        & \cite{carreira2021counterfactual} & QP & Constraint Search & CV, Tab. & No & Counterfactuals \\
        & \cite{kanamori2020dace} & MIP & Task-oriented Constraint & Tab. & No & Counterfactuals \\
        & \cite{albini2021influence} & Bayesian Opt. & Task-oriented Constraint & Tab. & Yes/No & Counterfactuals \\
        & \cite{looveren2021interpretable} & Spatial Opt. & Task-oriented Constraint & CV, Tab. & No & Counterfactuals \\
        & \cite{albini2022counterfactual} & Spatial Opt. & Shapley-based Constraint & Tab. & Yes & Counterfactuals \\
        % & \cite{lundberg2017unified} & Linear Model & Coefficient of Proxy Model & CV, Tab. & Yes & Feature Attribution\\
        & \cite{covert2021improving} & Linear Model &  Variance Reduction & Tab. & Yes & Feature Attribution \\
        & \cite{lundberg2020local} & Tree Model & Marginal Path Approximation & Tab. & Yes & Feature Attribution\\
        & \cite{karczmarz2021improved} & Tree Model & Banzhaf Values & Tab. & No & Feature Attribution\\
        & \cite{nid} & MLP Model & Examine from Model Weights & Tab. & No & Feature Interaction \\
        & \cite{pid} & MLP Model & Examine from Model Weights & CV, Tab. & No & Feature Interaction \\
        & \cite{shap_inter_idx} & Multilinear Extension & Weighted Least Squares Solver & NLP, Tab. & Yes & Feature Interaction \\
        & \cite{tsai2022faith} & Linear Model & Weighted Least Squares Solver & NLP & Yes & Feature Interaction \\
        \midrule\midrule
        Amtz. & \cite{cortx} & Self-supervised Learning & Contrastive Framework & CV, Tab. & Yes & Feature Attribution \\
        & \cite{jethani2021fastshap} & Supervised Learning & Sampling-generated Labels & CV, Tab. & Yes & Feature Attribution \\
        & \cite{chen2018learning} & Supervised Learning & Unified Feature Selection & CV, Tab. & No & Feature Attribution \\
        & \cite{schwab2019cxplain} & Supervised Learning & Unified Causality Relationship & CV, Tab., NLP & No & Feature Attribution \\
        & \cite{yang2021generative} & Adversarial Learning & GAN & CV, NLP & No & Counterfactuals \\
        & \cite{yang2021model} & Adversarial Learning & CGAN & Tab. & No & Counterfactuals \\
        & \cite{khorram2022cycle} & Adversarial Learning & GAN & CV & No & Counterfactuals \\
        & \cite{downs2020cruds} & Spatial Learning & Conditional Subspace VAE & Tab. & No & Counterfactuals \\
        & \cite{rodriguez2021beyond} & Spatial Learning & VAE & CV & No & Counterfactuals \\
       & \cite{antoran2020getting} & Spatial Learning & VAE & CV, Tab. & No & Counterfactuals \\
       & \cite{chen2021relax} & Reinforcement Learning & P-DQN Framework & Tab. & No & Counterfactuals \\
       & \cite{verma2022amortized} & Reinforcement Learning & PPO Algorithm & Tab. & No & Counterfactuals \\
    \bottomrule
    \end{tabular}}
\label{tab:deltalogodds}
\end{table*}  

\subsubsection{Feature Searching Optimization}
According to counterfactual optimization problems formulated in Equation~(\ref{eq:count}), the derivation process of counterfactual examples can be formulated as the contrasted optimization problems. Generally, traditional approaches such as integer programming are able to solve the problems and get the optimal solutions by adjusting the values of features. Nevertheless, the conventional optimization approaches are typically time-consuming due to their complex search steps. Some existing work even uses some kind of random or approximate search~\cite{karimi2020model,sharma2019certifai}, which are also computationally slow, particularly with high-dimensional instances.
The goal of feature searching optimization is to provide an efficient searching algorithm with linear or convex quadratic programming (QP) solvers to speed up the derivation of counterfactuals by changing the feature values. 

Prior works~\cite{mohammadi2021scaling,parmentier2021optimal,carreira2021counterfactual,russell2019efficient} focus on seeking efficient programming algorithms as the simulation of distance measurement function $\mathcal{L}(\cdot)$ in Equation~(\ref{eq:count}), providing the validity and proximity guarantees on counterfactual examples. One of the existing works~\cite{russell2019efficient} proposes a novel set of constraints, named mixed polytope, to incorporate with an integer programming solver to efficiently find coherent counterfactual explanations. The proposed mixed polytope can guarantee that the derived counterfactuals obtain a set of desiderata on diversity and closeness. 
Another work~\cite{mohammadi2021scaling} proposes efficient approaches to search for the nearest counterfactual explanation within a given interval in the input feature space, where the proposed searching ones can formulate the optimization problem with mixed-integer programming (MIP) solvers. The experimental results also prove a significant improvement in runtime efficiency in yielding MIP-based approaches for counterfactual generation.
Regarding the application of feature searching optimization,
a work~\cite{kaffes2021model} proposes several searching strategies that aim to efficiently extract concise explanations under constraints without access to the internal recommendation model. The key idea is to provide counterfactual explanations that are actionable and sparse, defined as minor changes to the user’s interaction logs for explaining recommendation outputs.

\subsection{Model-centric Acceleration} 
Model design is one of the essential factors that directly impact the performance of ML models, including speed and performance. Providing fitted task-oriented objectives and well-designed model structures can accelerate the training progress of ML models. In this manner, model-centric acceleration boosts the derivation speed of model explanation based on the model-wise adjustments, such as optimizing the objective functions or proposing new approximation models, which better estimate the explanation of the underlying prediction models with a faster speed.

\subsubsection{Optimization-driven Acceleration}  
Optimization-driven methods accelerate the explanation-deriving process by replacing it with relatively optimal training strategies. This line of work follows the philosophy that task-oriented training can solve efficient issues from the learning perspective, allowing it to converge faster. In this direction, we consider feature attribution tasks and counterfactual example tasks. Existing achievements focus on addressing the efficiency issues of counterfactual and feature attribution tasks. 

\begin{itemize}[leftmargin=*]
\item \textbf{Feature Attribution Tasks:}
Besides optimizing the training paradigm with the same objectives for faster derivation, some previous works reformulate the training objectives to accelerate the convergence speed of generating stable model explanations.
Existing work~\cite{mitchell2022sampling} proposes two simple and effective methods to accelerate the convergence of permutation sampling.
The first method is Kernel Herding~(KH) which adopts a dynamic process to select the buffer of permutations.
In each update step, a permutation is selected to minimize the Kendall correlation coefficient~(KCC)~\cite{kendall1938new} with the permutations in the buffer and maximize the KCC with the proposed efficient components.
In this manner, one more permutation is added to the buffer until the selected permutations reach the number limitation.
The second one is Orthogonal Spherical Codes (OSC), which reformulates the selection of permutations into orthogonal transformation in the hyper-sphere. To be concrete, OSC selects the correlated samples on the hyper-sphere from a basis of orthogonal vectors.
OSC can significantly accelerate the convergence since the KCC of the selected permutations can be explicitly bounded in the range of $[-\frac{1}{2}, +\frac{1}{2}]$, which is better than random sampling with $[-1, +1]$. However, the computational overhead of OSC is $O(n^2)$, where $n$ is the number of features.

\item \textbf{Counterfactual Example Tasks:}
As for counterfactual tasks, various of prior works~\cite{kanamori2020dace,dandl2020multi,albini2021influence,artelt2022efficient} proposes new loss functions to extract actions towards an interpretable counterfactual. \cite{looveren2021interpretable} offers a fast and model-agnostic method by incorporating the class prototypes in the objective function to guide the perturbations quickly to generate counterfactual explanations. The prototypes are beneficial to remove the computational bottlenecks caused by black-box prediction models. Another work, CounterfactualSHAP~\cite{albini2022counterfactual}, exploits "background information" with SHAP~\cite{lundberg2017unified} to guide on using feature attributions scores for rapidly generating counterfactual instances. The proposed framework comprises two new loss functions, the cost function and the action function, to enrich the counterfactual-ability of an approximated feature attribution from SHAP.

\end{itemize}

\subsubsection{Approximation-driven Acceleration}  
In this subsection, we introduce the approximation-driven methods according to feature attribution tasks and interaction detection tasks, respectively.

\begin{itemize}[leftmargin=*]
%\vspace{0.7mm}
% \noindent
\item \textbf{Feature Attribution Tasks:}
Approximation-driven methods propose white-box models to estimate the feature importance scores according to the ML predictions around the neighborhoods of the given inputs. Generally, the white-box models are self-interpretable and less complex, ensuring the explanation deviating process is fast and explainable. 
The white-box model does not have to work well globally, but it has to approximate the black-box model well in a small sampling neighborhood near the original input. Then the contribution score for each feature can be obtained by examining the parameters of the white-box model.
One existing work~\cite{covert2021improving} aims to rapidly estimate the unbiased feature importance scoring by approximating the optimal coefficients over the weighted linear regression.
Considering the prohibitively large ensembling times to achieve an unbiased explanation performance, this work proposes a variance reduction technique to speed up the approximation process of acquiring optimal coefficients as estimated unbiased Shapley values.
Unlike linear-based model approximation, some studies claim that the distribution of sampling neighborhood near the original input may be extremely non-linear, and linear-based model could lead to biased and ineffective estimations for model explanation~\cite{du2019techniques}. 
In this case, another group of works~\cite{karczmarz2021improved,lundberg2020local} takes advantage of non-linear models, such as tree-based models, to reduce the time complexity of estimating Shapley values. For instance, TreeSHAP~\cite{lundberg2020local} decreases the time complexity from $O(TL2^M)$ to $O(TLD^2)$ by reducing the number of trees (T) and maximum depth (D) of trees with $L$ leaves.

%\vspace{0.7mm}
% \noindent
\item \textbf{Feature Interaction Tasks:}
Besides the single feature attribution task that assumes each feature to be independent, some existing works focus on analyzing the attributions of feature interactions. However, the interaction candidacy is extremely huge. The number of interaction candidates is $2^N$ if there exist $N$ features. In the problem formulation in Equation~\ref{eq:inter_value}, traditional statistical post-hoc tests, such as ANOVA with Fisher's Least Significant Difference (LSD), need to conduct comprehensive tests through all interaction candidates, which is impractical for applying to large-scale problems. In this case, the goal of neural network-based interaction detection methods is to efficiently analyze the statistical interactions of Equation~\ref{eq:inter_value}, which are highly related to the results of prediction models. One of the existing works, NIT \cite{nit}, first demonstrates that any interacting features must follow strongly weighted connections to a common hidden unit before reaching the final output layer. Based on this observation,
NID~\cite{nid} detects interactions from weights of learned neural networks by examining the interacting paths between the weight of the first layer and those from subsequent layers.
PID~\cite{pid} further explored the idea of interacting paths by extending the theory of persistent homology to the interaction detection problem. Although these methods claim to generate a post-hoc global-wise explanation, it is still available to generate locally instance-wise explanation~\cite{pid} with interaction importance scores. 
Regarding the application of interaction detection methods,
GLIDER \cite{Tsang2020Feature} utilizes gradient interaction detection methods to efficiently build synthetic crossing features for each detected group of interacting features to improve the recommendation performance.

Another aspect is the acceleration of Shapley-based interaction scores. Following the definition of Shapley interaction index~\cite{grabisch1999axiomatic}, the Shapley interaction scores $\phi_I (f, \mathcal{U})$ of targeted feature interaction $I \subseteq \mathcal{U}$ can be defined as:
\begin{equation}
    \vspace{0.1cm}
    \notag \!\!\phi_I (f, \mathcal{U}) \!=\!\! \sum_{\boldsymbol{S} \subseteq \mathcal{U} \setminus I} \!\! \binom{M \!-\!\! 1}{|\boldsymbol{S}| \!\!- |I|\!+\!\! 1| }^{-1} \!\!\!\!\! (-1)^{|I| - |L|} \sum_{L \subseteq I} f(L \cup \boldsymbol{S}),
    \label{eq:inter_shapley_value}
    \vspace{0.1cm}
\end{equation}
where $L$ represents the potential of feature sub-interactions happening in targeted feature interaction $I$. However, it is extremely time-consuming to calculate the exact Shapley interaction scores. The time complexity $O(2^{M+|I|})$ is even higher than the exact calculation of Shapley values for individual features, which is implausible to apply to any real-world system. 
In this manner, some prior works focus on approximating the interaction scores on top of Shapley interaction scores. 
One of the works, Faith-SHAP \cite{tsai2022faith}, aims to extend Shapley values from individual feature importance scores to interaction importance scores. 
The key idea is to interpret Shapley values as coefficients of the most faithful linear approximation to the pseudo-Boolean coalition game value function, where the process of leveraging the weighted linear approximation can lead to efficient explanation derivation of feature interaction.
Another work, Shapley-Taylor indices~\cite{shap_inter_idx}, detects feature interactions by expanding the Taylor series of the multi-linear extension with the set-theoretic model behavior. Under the samples of interaction permutations by random process over orderings of features, the derivation process can be accelerated through the approximation. By exploiting the concept of second-order Shapley-Taylor indices, \cite{hamilton2021axiomatic} extend traditional non-amortized methods, such as GradCAM, LIME, and SHAP, to extract pairwise correspondences between images from trained opaque-box models.
The other work~\cite{tsang2020does} detects feature interaction based on Taylor expansion by interpreting the importance of interactions as mixed partial derivatives, which achieve a runtime-efficiency process.
\end{itemize}

\section{Amortized Acceleration} 
After understanding the non-amortized ways to accelerate post-hoc local explanation, we turn to introduce how amortized acceleration affects the efficiency of deriving model explanation. Unlike non-amortized ones, amortized methods require only one explanation model to generate the model explanation among all instances, which speeds up the progress with its fast inference phase of a unified explainer. In this section, we introduce three different mechanisms of amortized acceleration methods on both feature attribution and counterfactual tasks.

\subsection{Predictive-driven Method}
The predictive-driven methods maintain a unified DNN explainer to generate a fast model explanation among each data instance. The explanation can be generated via a single feed-forward process of the DNN explainer, providing real-time estimation on feature attribution tasks. Compared to the existing traditional XAI methods, the advantages of the predictive model mainly lie in two folds: (1) faster explanation generation; and (2) more robust explanation derivation. Generally, existing works attempt to learn the overall explanation distribution using two lines of methodologies, which are attribution-based approaches~\cite{wang2021shapley,jethani2021fastshap,covert2022learning,covert2022learning,cortx} and prior-knowledge-based approaches~\cite{chen2018learning, dabkowski2017real, kanehira2019learning, schwab2019cxplain, fu2021differentiated, konstantinov2022attention, schwarzenberg2021efficient, hesse2021fast,jethani2021have}. 

The first line of work employs the DNN explainers to simulate a given approximated Shapley distribution for generating explanation results. One of the representative works, FastSHAP~\cite{jethani2021fastshap}, exploits a DNN model to capture the Shapley distribution among training instances for efficient real-time explanations, which is a supervised paradigm learned with approximated Shapley value labels. Although the approximate attribution labels are rapid, degradation has been shown to affect the performance of explanation broadly~\cite{cortx}. Another work, CoRTX~\cite{cortx}, proposes an unsupervised learning paradigm, which can significantly reduce the dependency on the Shapley labels and accelerate the derivation progress. The unsupervised CoRTX benefits the explanation tasks by exploiting a contrastive framework for generating latent explanations. After that, CoRTX fine-tunes the latent explanations with extremely few-shot labels to get the final model explanation. 

The second line of work assumes the specific pre-defined causal relationship or feature distributions, as well as formulates the explainer learning process based on the given assumptions. One of the works, L2X~\cite{chen2018learning}, provides a feature masking generator for real-time feature selection. The training process of the mask generator is under the constraint from the given predicted label distribution of masked imputing instances. In addition to the previous definition of data distribution, CXPlain~\cite{schwab2019cxplain} train an explanation model by using a causal objective function that follows the definition of Granger causality~\cite{granger1969investigating}. To guarantee efficacy, CXPlain provides the uncertainty estimations for feature importance scores that are strongly correlated with the efficacy of the provided importance scores on previously unseen test data. 

In this manner, both the first line and second line of work exploit DNN models to learn the unified distribution of model explanation for providing the explanation derivation with pre-trained explainers. Due to the inference phase of well-trained explainers, predictive models can provide real-time model explanations.

\subsection{Generative-driven Method}
The goal of generative methods is to learn the unified counterfactual rules by utilizing generative models from raw data instances such as textual and image data. The derivation process can be effective and efficient by exploiting generative models conditioned with a counterfactual universe. Unlike traditional local-wise counterfactual methods that modify instances in the data space, generative models focus on constructing the attribute-informed latent space, guiding the generative models to fit the counterfactual distribution of multiple query data instances. 
Existing work proposes their DNN frameworks based on two training paradigms: adversarial-learning-based~\cite{yang2021model,yang2021generative,singla2019explanation,singla2021explaining,antoran2020getting,khorram2022cycle} and spatial-learning-based~\cite{pawelczyk2020learning,joshi2019towards,rodriguez2021beyond,guyomard2022vcnet,downs2020cruds,mahajan2019preserving,joshi2018xgems,pawelczyk2020learning,ma2022clear}, which allows counterfactual examples to be rapidly generated by reusing same models among multiple queried data instances. 

The first aspect is adversarial-learning-based ones, which utilize adversarial learning to guarantee the effectiveness of derived counterfactuals. A representative adversarial-based work, MCS~\cite{yang2021model}, constructs a model-based synthesizer by using a conditional generative adversarial network (CGAN)~\cite{mirza2014conditional} to capture counterfactual information faithfully. By incorporating model inductive bias, MCS can accurately exploit the causal dependence of attributes, which attempts to ensure the design correctness of the generative models through the causality identification process. The second aspect is spatial-learning-based approaches, which adopt generative models to build up the algorithmic recourse frameworks by reformulating the counterfactual latent space. One of the existing works, DiVE~\cite{rodriguez2021beyond}, is built upon the VAE~\cite{kingma2013auto}-based structure with counterfactual constraints, allowing the deviation process to be efficiently completed via single forward passing. This work aims to learn the disentangled latent space by leveraging the Fisher information matrix of the underlying prediction ML models. With the learned representation, the generated counterfactual explanation is enforced to be proximity, sparsity, and diversity. However, compared to non-amortized counterfactual methods, there is a trade-off between explanation efficacy and efficiency in generative-driven strategies, making it a remaining challenge to balance these two goals in the training phase. 

\subsection{Reinforcement Agent}
Reinforcement agents aim to reformulate the counterfactual explanation problem into sequential decision-making progress, where they utilize reinforcement learning to find the optimal counterfactual instances. Owing to the fast inference phase of reinforcement agents, the explanation deviating process is extremely fast and can generate multiple instances simultaneously.
Prior work, RELAX~\cite{chen2021relax}, has produced model-agnostic counterfactual examples using the P-DQN framework. The derivation process is reformulated as a Markov decision process (MDP) with hybrid discrete-continuous actions to ensure the sparsity and proximity of the generated counterfactuals. 
Experimental results have shown that these agents lead to significant improvements in both efficacy and efficiency. Another work~\cite{verma2022amortized} propose 
a stochastic-control-based approach that uses the proximal policy optimization (PPO) algorithm to generate sequential algorithmic recourse.
The derivation process allows the data instance to move stochastically and sequentially across the intermediate states to reach the final generated examples, which are treated as the generated counterfactuals. 
Following the practical guidance of algorithmic recourse, the framework translates the algorithmic recourse problem into an MDP with the constraints of discrete state space and discrete action space. In particular, the design of state space ensures actionability and action space fulfills the sparsity. The evaluation results indicate the successful generation of sequential counterfactual instances that meet other recourse desiderata.

\section{Discussion of the Limitations}
% Discuss the current methods (comparison w/ non-amt \& amt ) \\
% Model aspect: \\ 
%    5.1 - Training efficient (training speed / Label usage) \\
%    5.2 - Deployment efficient (model size)  [gradient-based vs RTX] \\
%    5.3 - Explanation evaluation --> good or bad 
%    5.4 - How to choose: amortized or non-amortized // (ANOVA vs. NID)
%        - Trade-off between efficacy and efficiency: Experiment results on executing speed \\

\begin{table*}[ht]
\centering
\caption{Experiment results of feature attribution task on CIFAR-10 dataset from two previous works~\cite{jethani2021fastshap,cortx}. "Millisecond" (ms) is the measurement unit of deployment time.}
% "Second" is the unit of measuring training time per data instance, while 
\setlength{\tabcolsep}{5pt}{
\begin{tabular}{l | ccc ccc ccc ccc ccc ccc ccc ccc ccc ccc|}
\toprule
CIFAR-10 Dataset & CoRTX~\cite{cortx} & FastSHAP~\cite{jethani2021fastshap} & SHAP~\cite{lundberg2017unified} & IG~\cite{sundararajan2017axiomatic} & SmoothGrad~\cite{smilkov2017smoothgrad} & GradCAM~\cite{selvaraju2017grad} & DeepSHAP~\cite{lundberg2017unified} \\
\midrule\midrule 
% Training Time (sec.) & & 41.6142 & --- & --- & --- & --- & --- \\
Deployment Time (ms) & 0.4 & 0.4 & 27221.4 & 54.6 & 60.0 & 22.8 & 323.4 \\
% Exclusion AUC & \textbf{0.373} $\pm$ 0.011 & 0.420 $\pm$ 0.015 & 0.395 $\pm$ 0.012 & 0.559 $\pm$ 0.012 & 0.471 $\pm$ 0.011 & 0.563 $\pm$ 0.012 & 0.554 $\pm$ 0.012 \\
% Inclusion AUC & \textbf{0.774} $\pm$ 0.012 & \textbf{0.782} $\pm$ 0.015 & 0.769 $\pm$ 0.012 & 0.740 $\pm$ 0.012 & 0.738 $\pm$ 0.012 & 0.741 $\pm$ 0.012 & 0.742 $\pm$ 0.012\\
\bottomrule
\end{tabular}}
\vspace{2mm}
\label{tab:traineff}%
\end{table*}%

\begin{table*}[ht]
\centering
\caption{Experiment of feature attribution tasks on Adult dataset from two previous works~\cite{wang2022accelerating,cortx}. We utilize throughput to measure the deployment time.  Higher throughput indicates higher efficiency of the explanation process.}
% "Second" is the unit of measuring training time per data instance, while 
\setlength{\tabcolsep}{5pt}{
\begin{tabular}{l | ccc ccc ccc ccc ccc ccc ccc ccc ccc ccc|}
\toprule
Adult Dataset & CoRTX~\cite{cortx} & FastSHAP~\cite{jethani2021fastshap} & SHEAR~\cite{wang2022accelerating} & SHAP~\cite{lundberg2017unified} & Permutation Sampling~\cite{mitchell2021sampling} \\
\midrule\midrule 
% Training Time (sec.) & & 41.6142 & --- & --- & --- & --- & --- \\
Deployment Time (Throughput) & 6202.85 & 6202.85 & 71.064 & 1.0940 & 22.077\\
% \makecell[c]{Semi-Supervised \\ Learning}}
% Explanation Acc. ($\boldsymbol\ell_2\text{-error}$) & 0.0199 & 0.00723 & & 0.0174 & 0.0175 \\
\bottomrule
\end{tabular}}
\vspace{2mm}
\label{tab:trainetab}%
\end{table*}%

% We briefly introduce the explanation models according to three different aspects, which are training efficiency, deployment efficiency, explanation evaluation, and the tips to adopt non-amortized acceleration and amortized acceleration methods in certain scenarios.
We briefly introduce the limitations of efficient XAI we have surveyed from the training phase, deployment phase, and using scenarios. Then we present suggestions for choosing efficient XAI methods that might be more suitable to match the using scenarios for users.

\subsection{Limitation on Training Phase}
The learning strategies of non-amortized and amortized methods have been a bottleneck in making XAI local methods efficient. Current efficient non-amortized explanations are usually given in reducing queried features and providing advanced efficient-oriented algorithms. Without the training-testing process, non-amortized methods can neglect to prepare the pre-trained explainers. In contrast, amortized methods need to train a unified model before deriving a model explanation, causing additional computational resources to an explainer model in advance. In addition, the training speed of amortized ones highly depends on the explanation data scale, which can be referred to as the usage amount of explanation label in the training phase of supervised learning paradigm~\cite{jethani2021fastshap,yang2021generative}, and the fine-tuning stage of self-supervised learning paradigm~\cite{cortx}. This means that the amortized method moves the heavy computational complexity to the training phase, leading to a fast derivation via its single-forward inference process. There exists a trade-off between training speed and explanation performance in amortized methods on feature attribution and counterfactual tasks. 
In general, the trade-off between training speed and explanation performance does not have a significant impact in real-time online service, as amortized explanation models are typically pre-trained. However, if a production system requires online learning~\cite{mcmahan2013ad}, the long training time required for amortized methods could be a barrier to deployment.

\subsection{Limitation on Deployment Phase}
There are several metrics that can be used to evaluate the efficiency of non-amortized and amortized acceleration methods in the deployment phase. One of the commonly used metrics is algorithmic throughput~\cite{wang2022accelerating,cortx,teich2018plaster}.
Specifically, the throughput is calculated by $\frac{N_{\text{test}}}{t_{\text{total}}}$, where $N_{\text{test}}$ and $t_{\text{total}}$ denote the testing instance number and the overall time consumption of explanation derivation, respectively. In this case, higher throughput indicates higher efficiency of the explanation process. Table~\ref{tab:trainetab} indicates the execution time of the deployment phase. The experimental results show that the efficient amortized methods (e.g., CoRTX and FastSHAP) obtain significantly larger algorithmic throughput than efficient non-amortized methods (e.g., SHEAR). In addition, efficient non-amortized methods significantly improve the deployment time compared to traditional non-amortized methods.
Instead of fixing the time to count the executed instances (e.g., algorithm throughput), some other works~\cite{jethani2021fastshap} use the metric by fixing the tested instances and reckoning the execution time.

Despite the significant improvement in the efficiency of non-amortized ones, there is still a considerable gap in the deployment of real-time online services. Compared to amortized methods that only require an extremely short time to derive explanations, traditional non-amortized ones, such as SHAP~\cite{lundberg2017unified}, can take up to 200x longer to complete the explanation process~\cite{jethani2021fastshap,cortx}. While non-amortized methods are more accurate than amortized ones, their slower execution time makes them impractical for real-time applications.
Some gradient-based methods, such as \cite{sundararajan2017axiomatic, smilkov2017smoothgrad}, are important explanation methods that yield a relatively faster explanation than other traditional non-amortized methods. However, gradient-based methods still need extra time to conduct the sampling process on each data instance, slowing the execution time while generating the instance-wise model explanation. 
The explanation derivation time highly depends on the scalability of sampling and testing instances. As a result, 
%amortized models are still significantly faster than gradient-based methods, making them 
gradient-based methods are inadequate for deployment in real-time systems. In Table~\ref{tab:traineff}, we conduct an experiment on CIFAR-10 dataset to observe the efficiency of explanation methods. The experimental results reveal the execution time on the inference phase per image. The results show that amortized-based methods are much faster than other efficient non-amortized methods (e.g., gradient-based methods), indicating that amortized ones are still the best candidate to deploy on real-time systems.

\subsection{Limitation on Using Scenario}
%    5.3 - Explanation evaluation --> good or bad 
%    5.4 - How to choose: amortized or non-amortized // (ANOVA vs. NID)
%        - Trade-off between efficacy and efficiency: Experiment results on executing speed \\
% The trade-off between efficacy and efficiency makes it difficult to choose between efficient non-amortized and efficient amortized acceleration methods. 
The trade-off between efficacy and efficiency brings the limitation to choosing efficient XAI methods for certain using scenarios appropriately. One reason is that selecting the best algorithm for a given scenario necessitates balancing the two factors. This is because a highly effective but inefficient algorithm may be impractical for large-scale applications, whereas a highly efficient but ineffective algorithm may not provide useful solutions. One prior work~\cite{cortx} designs a flexible data proportional parameter on the amortized method, providing an extra degree of freedom to coordinate the balance of efficacy and efficiency based on the users' specific requirements of the problem at hand. Another example is interaction detection tasks. Users may choose different interaction detection techniques based on the prediction models. While explaining the impact of feature interaction toward multi-layer perception (MLP), NID and PID can provide fast and accurate explanations via detecting interactions of variable order, which are extremely time-consuming by using traditional statistical tests (e.g., four-way ANOVA). 
When it comes to analyzing the prediction models that are not in the zone of MLP-based models (e.g., probit models and logistic models), NID and PID are not well-suited, whereas traditional statistical tests, such as~\cite{ai2003interaction}, are able to fit in this situation. Therefore, the decision to use amortized or non-amortized acceleration methods should be based on the specific needs of the user, including the type of prediction model being targeted and the tolerance for computation latency and performance.

\section{Research Challenges}
% (non-amt \& amt) \\
% Deployment: \\
%     - Software aspects: \\
%         - Centralized \& decentralized \\
%     - Hardware aspects \\
%         - Limited computational resources \\
%         - Distributed system \& edge device \\
% Human-centric issues:\\
%     - Privacy \\
%     - Fairness \\
%     - Security \\
% Move to challenges
% Human aspect: \\
%     - How fast is enough to fit the scenario \\
%     - Healthcare 

Despite recent advances in efficient XAI, there are still several urgent challenges, particularly in the deployment of efficient explanation methods and the issues of their trustworthiness raised by stakeholders and regulations. These challenges need to be addressed in order to successfully deploy XAI in real-world applications.

\subsection{Deployment on Efficient XAI}
The first research challenge lies on the deployment side. We present the deployment challenges for non-amortized acceleration and amortized acceleration methods in two parts: software and hardware. On the software side, we explore centralized and decentralized training. On the hardware side, we discuss the limitations of different devices and the impact on the deployment of acceleration methods. However, only limited previous works~\cite{barcena2022fed} discuss the topic.

\subsubsection{Non-amortized Acceleration Methods}
The efficiency improvement from non-amortized acceleration methods makes it possible to deploy on real-world applications. Non-amortized ones obtain the advantage of deploying with decentralized training since each explainer is independent, which opens a big map on real-world deployment. Despite its natural advantage, the decentralized training paradigm may face a performance drop, since we cannot manage the training resource individually to meet Pareto efficiency.
It is also difficult to coordinate conflicts consistently made by different local explainers when encountering similar data instances but with a different model explanation. One of the possible ways is to use federated learning~\cite{aledhari2020federated} with a centralized working node.
On the hardware side, efficient non-amortized methods are possibly applicable to the edge-computing paradigm. One of the main reasons is that the requirement for computational resources is relatively lower than traditional XAI local methods. Different from cloud computing platforms, edge computing has the advantages of low latency of network congestion but limited data storage and computing resources, which efficient non-amortized ones can partially handle. Additionally, it is essential to establish a standardized experimental setup for analyzing both i.i.d. and non-i.i.d. data distributions among edge clients. Finding the right balance between the software and hardware is still an open problem for deploying efficient non-amortized methods to real-world applications.

\subsubsection{Amortized Acceleration Methods}
Unlike non-amortized acceleration explainers that are separable and independent, amortized acceleration methods aim to generate a unified explainer for real-time derivation, which requires a sufficient explanation dataset for centralized training. This may cause risks in collecting private user data from different devices, including data leakage or attack. One of the challenges in amortized ones is how to efficiently and effectively train them under the decentralized learning paradigm, which prevents the data gathering requests to the clients. 
In terms of the hardware side, amortized acceleration explainers can provide real-time services on both edge devices and online service platforms. Based on the current state-of-the-art efficient amortized ones, the computational requirements are much less than non-amortized ones, making it applicable to deploy on edge-computing devices with limited computational resources. However, a pre-trained amortized explainer may be large and complex, which requires substantial storage and computing capacity and obtains a long training process.
In this way, the distributed training paradigm~\cite{jia2019beyond} can be one of the possible ways to cope with this difficulty. Generally, data parallelism and model parallelism are the two main types of distributed training. From a data parallelism perspective, a single worker device needs without large memory usage and disk storage, contributing to faster training time. However, there are times when the explanation models are too large to fit in a single worker device, and thus model parallelism is here to accelerate the training process from another perspective.
As for the model parallelism, the explanation model itself is split into several parts that are trained simultaneously across different worker devices. Throughout sharing the model weights from different devices, the final explainer can be provided with a shorter training time with parallel architectures. Nevertheless, the design of an efficient training paradigm on amortized methods still remains an open problem in generating faithful and accurate model explanations. 

\subsection{Trustworthy Issues in Efficient XAI}
The second research challenge involves issues of three parts: privacy, fairness, and security. Careful consideration must be given to these potential issues in order to ensure the trustworthy and responsible deployment of amortized acceleration methods of efficient XAI. We introduce the human-centric challenges below for non-amortized acceleration methods and amortized acceleration methods. 

\subsubsection{Issues on Non-amortized Acceleration Methods}
Non-amortized acceleration methods in XAI could be subject to fairness and security issues. One of the issues is the fairness problem caused by sensitive attributes in the original prediction model. No matter in attribution or influential example task, a fair situation of model explanation should remain unchanged when sensitive attributes, such as race and gender in the criminal dataset, are considered~\cite{kusner2017counterfactual}.
Typically, sensitive attributions are the essential features of the biased prediction model, and data-centric methods, for example, can then easily pick up those sensitive attributions as essential features to further explainer learning. This can result in the inclusion of biased information from the prediction model, which will be further emphasized after the explainer learning, leading to fairness issues. Another one is the security issue in non-amortized ones. 
In non-amortized settings, each individual user obtains their own explainer, which heavily overfits the data from a specific user. Model-centric acceleration methods can easily suffer from the data poison attack~\cite{lin2021ml} since efficient optimization is sensitive to the adjustment of data values. In addition, data-centric methods are also vulnerable to data poison attacks, where the search optimization process can be significantly affected if the feature values are not of good quality. Thus, it is essential to carefully consider and address these issues in developing non-amortized acceleration methods for XAI.

\subsubsection{Issues on Amortized Acceleration Methods}
The use of amortized acceleration methods also poses an increased risk of encountering privacy, fairness, and security issues, as these methods rely on additional DNN models to learn the explanation distribution from individual users.
One risk is the potential for privacy violations. Amortized acceleration methods require a large amount of private data to establish, since the targeted features to explain are usually the personal features of individual users. In an amortized setting, all users rely on the same explainer, which means that private information can be easily shared through the public access of an explainer without their consent or knowledge. This raises additional concerns about the security and privacy of sensitive information. Another issue faced by amortized ones is fairness issues led by the biased training patterns in the DNN-based explainer. This is the challenge for most DNN-based models trained with sensitive attributes~\cite{du2020fairness}. The learning paradigm may tend to learn the spurious relevance between specific sensitive attributes and explanation results, causing algorithmic discrimination. The fairness issues should be carefully emphasized in the future direction of efficient XAI.  
Last but not least, one of the important issues that need to be awarded for amortized ones is security. Security measures are designed to protect user data from being hacked or stolen. However, these measures can be easily circumvented if the explainer is encoded with whole batches of user data, such as using the techniques of model inversion attack~\cite{chakraborty2018adversarial}. The attack can recover personal information from API access to amortized explanation models. This emphasizes the importance of security issues in amortized explanation models.

\subsection{Towards Human-aware Selection}
Efficient XAI provides fast and accurate model explanations to users, enabling them to trust the decision making of prediction models. An open question can be raised here: How can we effectively choose a type of explainer in order to provide a satisfactory explanation for certain scenarios?
It is necessary to carefully choose the desired types of explainers according to the given tasks. For example, autopilot systems require the explanation derivation time to be extremely short for real-time judgments, whereas medical diagnosis cases are relatively flexible in the speed of generating model explanations. One criterion to consider when choosing an explainer is the tolerance of latency in deriving a model explanation. 
For example, in real-time bidding systems, users require real-time model explanations to support the real-time decisions made by bidding systems. However, efficient non-amortized methods may not be fast enough to meet the tight deviation time constraints.
Efficient amortized methods are able to meet the real-time requirements, as they only obtain a single forward pass during the inference phase.
Another example is online streaming services, where users are expected to obtain fast and personalized explanations. 
Efficient non-amortized methods are particularly useful in this context because they typically provide personalized explanations.
In this case, efficient non-amortized methods are more suitable to fit the requirements than amortized methods, as they provide each user with its own unique explainer. As a result, the usage scenarios of the efficient approaches are one of the essential criteria leading to an effective derivation process.

% [How to choose amortized or non-amortized: personalized issue]
% One open question can be raised here: How efficient is enough in order to require a satisfactory explanation result?
% To better cope with this challenge, there are several kinds of mechanisms in the HCI domain that can qualitatively and quantitatively analyze the user's preference and experience on a specific explanation scenario, such as semi-structured interview~\cite{horton2004qualitative} and user experience study~\cite{bargas2011old}. However, considering the necessity of efficiency, there is no prior study concentrated on acceptable execution time by analyzing through either user interviews or studies. This is one of the current deployment challenges that efficient XAI may encounter.

\section{Conclusion}
Efficient XAI is a rapidly evolving research area with significant demand from practical applications. This work provides a clear taxonomy and a comprehensive overview of existing techniques to aid practitioners and researchers in selecting the most suitable efficient algorithms for their specific needs. Examining the perspective of efficiency may be beneficial to the community in better understanding the limitations of existing XAI methods. 
Despite the progress in efficient XAI, there are still significant challenges that require future solutions to further emphasize the importance of efficiency. We hope that this work will serve as a valuable resource for both newcomers and professionals who are interested in the broad field of efficient XAI.

% \section*{Acknowledgment}

\ifCLASSOPTIONcaptionsoff
  \newpage
\fi

\bibliographystyle{IEEEtran}
\bibliography{paper}

\begin{IEEEbiography}
[{\includegraphics[width=1in,height=1.25in,clip,keepaspectratio]{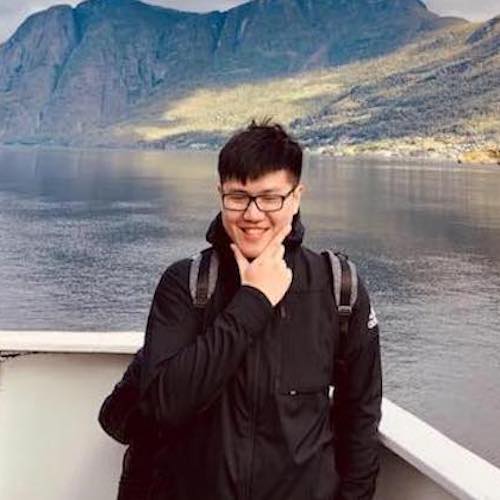}}]
{Yu-Neng Chuang}
is a second-year Ph.D. student in the Computer Science Department at Rice University. His research interest lies in trustworthy machine learning, including explainable artificial intelligence, machine learning fairness, and recommender systems. Currently, he is focusing on the efficiency of interpretable machine learning and fair modeling. Prior to Rice, his previous research also involves recommender systems via modeling user behaviors and sparse labeling of textual data. Yu-Neng received his B.S. and M.S. degrees in Mathematics and Computer Science from National Chengchi University, respectively in 2017 and 2020. 
\end{IEEEbiography}

\begin{IEEEbiography}
[{\includegraphics[width=1in,height=1.25in,clip,keepaspectratio]{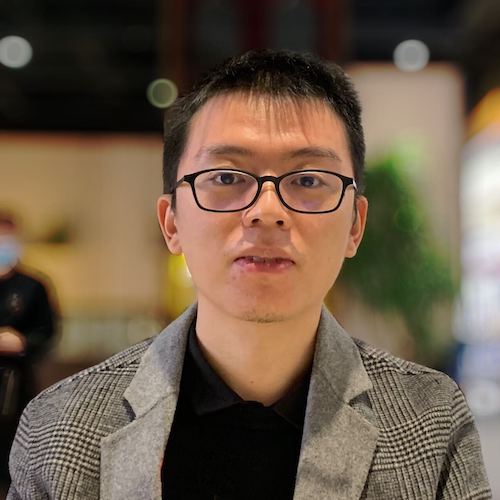}}]
{Guanchu Wang}
Guanchu Wang received the BS and MS degrees in electrical engineering and information science from Dalian University of Technology and University of Science and Technology of China, respectively. He is currently a Ph.D. student in the department of computer science at Rice university. His research focuses on efficient and interpretable machine learning. 
\end{IEEEbiography}

\begin{IEEEbiography}
[{\includegraphics[width=1in,height=1.25in,clip,keepaspectratio]{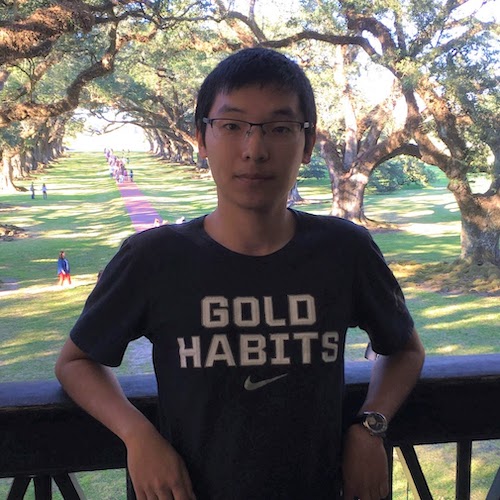}}]
{Fan Yang}
Fan Yang is currently a final-year Ph.D. student in the Computer Science Department at Rice University. His research interests generally lie in the area of eXplainable Artificial Intelligence (XAI), with a major focus on model interpretation techniques, counterfactual explanation, and machine learning fairness. He is also interested in XAI-related downstream application, as well as its correlative intersections with Natural Language Processing and Human-Computer Interaction. Prior to Rice, Fan had research experiences on wireless communication and networking. Fan received his M.S. and B.S. degree in Xidian University, respectively in 2016 and 2013. 
\end{IEEEbiography}

\begin{IEEEbiography}
[{\includegraphics[width=1in,height=1.25in,clip,keepaspectratio]{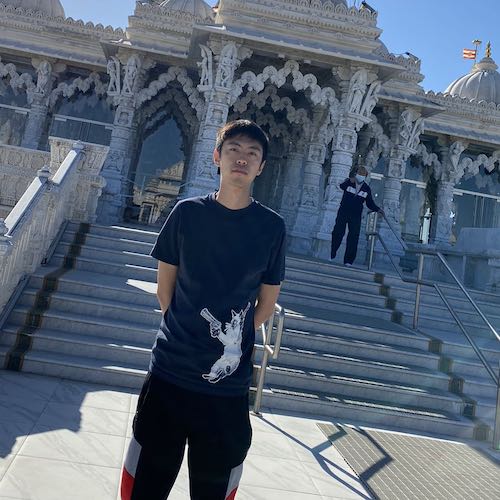}}]
{Zirui Liu}
Zirui Liu received BS and MS degrees in electrical engineering both from Harbin Institute of Technology. He is currently a Ph.D. student in the Department
of Computer Science at Rice University. His research interests include large-scale machine learning and graph neural networks.
\end{IEEEbiography}

\begin{IEEEbiography}
[{\includegraphics[width=1in,height=1.25in,clip,keepaspectratio]{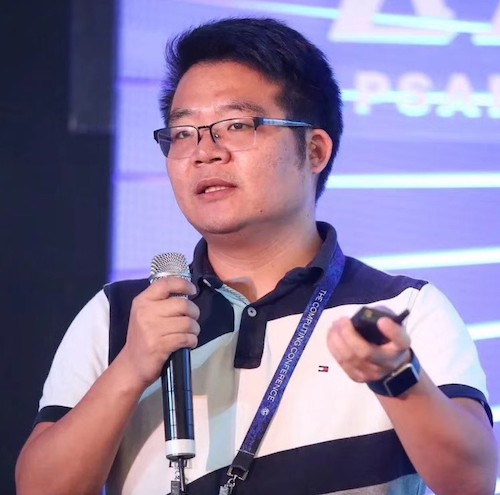}}]
{Xuanting Cai} is a software engineer at Meta Platforms, Incorporated. Xuanting Cai earned his Ph.D. in Mathematics from Louisiana State University and B.S. in Mathematics and Economics from Peking University. Before joining Meta, he worked at Alibaba.com and Google as a software engineer.
\end{IEEEbiography}

\begin{IEEEbiography}
[{\includegraphics[width=1in,height=1.25in,clip,keepaspectratio]{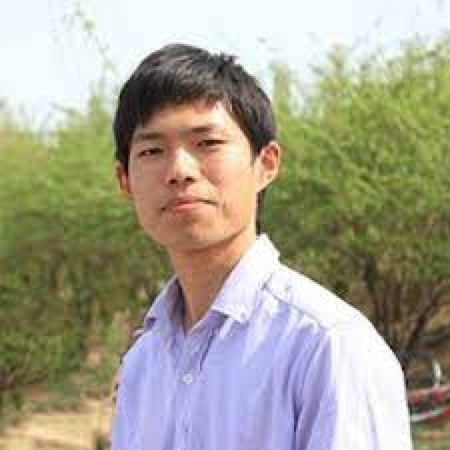}}]
{Mengnan Du}
is an Assistant Professor in the Department of Data Science, New Jersey Institute of Technology (NJIT). Mengnan Du earned his Ph.D. in Computer Science from Texas A\&M University. He has previously worked/interned with Microsoft Research (MSR), Adobe Research, Intel, Baidu Research, Baidu Search Science and JD Explore Academy. His research covers a wide range of trustworthy machine learning topics, such as model explainability, fairness, and robustness. He has had more than 40 papers published in prestigious venues such as NeurIPS, AAAI, KDD, WWW, ICLR, and ICML. He received over 2,300 citations with an H-index of 16. 
\end{IEEEbiography}

\begin{IEEEbiography}
[{\includegraphics[width=1in,height=1.25in,clip,keepaspectratio]{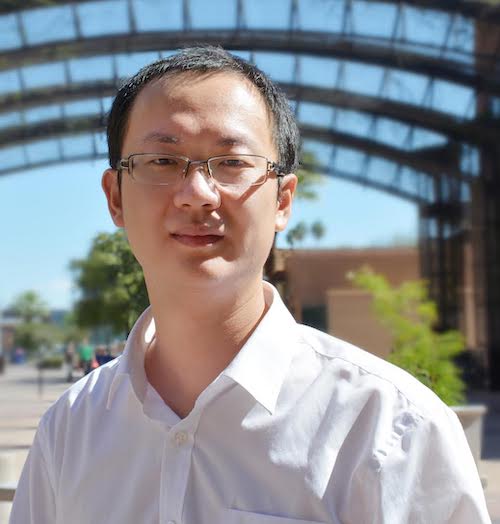}}]
{Xia “Ben” Hu} is an Associate Professor at Rice University in the Department of Computer Science. Dr. Hu has published over 100 papers in several major academic venues, including NeurIPS, ICLR, KDD, WWW, IJCAI, AAAI, etc. An open-source package developed by his group, namely AutoKeras, has become the most used automated deep learning system on Github (with over 8,000 stars and 1,000 forks). His papers have received several Best Paper (Candidate) awards from venues such as ICML, WWW, WSDM, ICDM, AMIA and INFORMS. He is the recipient of NSF CAREER Award and ACM SIGKDD Rising Star Award. His work has been cited more than 16,000 times.
\end{IEEEbiography}

\end{document}